\documentclass[]{spie}  

\usepackage{amsmath,amsfonts,amssymb}
\usepackage{graphicx}
\usepackage[colorlinks=true, allcolors=blue]{hyperref}
\title{Automated Plaque Detection and Agatston Score Estimation on Non-Contrast CT Scans: A Multicenter Study}

\author[a]{Andrew M. Nguyen}
\author[a]{Jianfei Liu}
\author[a]{Tejas Sudharshan Mathai}
\author[b]{\\ Peter C. Grayson}
\author[a]{Ronald M. Summers}
\affil[a]{Imaging Biomarkers and Computer-Aided Diagnosis Laboratory, Radiology and Imaging
Sciences, National Institutes of Health Clinical Center, Bethesda, MD 20892-1182, USA
}
\affil[b]{National Institute of Arthritis and Musculoskeletal and Skin Diseases, National Institutes of Health, Bethesda, Maryland
}

\authorinfo{Further author information: (Send correspondence to J.L.)\\J.L.: E-mail: jianfei.liu@nih.gov}

\begin{document} 
\maketitle

\begin{abstract}
Coronary artery calcification (CAC) is a strong and independent predictor of cardiovascular disease (CVD). However, manual assessment of CAC often requires radiological expertise, time, and invasive imaging techniques. The purpose of this multicenter study is to validate an automated cardiac plaque detection model using a 3D multiclass nnU-Net for gated and non-gated non-contrast chest CT volumes. CT scans were performed at three tertiary care hospitals and collected as three datasets, respectively. Heart, aorta, and lung segmentations were determined using TotalSegmentator, while plaques in the coronary arteries and heart valves were manually labeled for 801 volumes. In this work we demonstrate how the nnU-Net semantic segmentation pipeline may be adapted to detect plaques in the coronary arteries and valves. With a linear correction, nnU-Net deep learning methods may also accurately estimate Agatston scores on chest non-contrast CT scans. Compared to manual Agatson scoring, automated Agatston scoring indicated a slope of the linear regression of 0.841 with an intercept of +16 HU (R² = 0.97). These results are an improvement over previous work assessing automated Agatston score computation in non-gated CT scans.

\end{abstract}

\keywords{CT, non-gated, coronary artery calcification, deep learning, detection}

\section{Purpose}

\label{sec:Purpose}  

Coronary artery calcification (CAC) is a strong and independent predictor of cardiovascular disease (CVD)\cite{Summers2021}. Inflammation, necrosis, fibrosis, or calcification can lead to atherosclerosis, the build-up of plaque, which may obstruct blood flow from the heart and manifest in CAC and eventually cardiovascular disease (CVD)\cite{Severino2020,Bentzon2014}. Plaques that rupture or erode over time may induce life-threatening coronary thrombosis, acute coronary syndrome (ACS), and myocardial infarction (MI).

Manual assessment of CAC often requires radiological expertise, time, and invasive imaging techniques\cite{Jiang2020}. The CAC burden is typically quantitatively assessed using the Agatston score. This score is determined by summing the product of each plaque's calcium area and its attenuation above 130 Hounsfield Units (HU), thereby providing a measure of plaque burden\cite{Greenland2018}. The Agatston score can then be used to stratify risk in clinical settings, such as for diabetes and severe hypercholesterolemia management\cite{Noguchi2018,Nasir2021}.

Beyond CAC, heart valve calcification (VC) in the aortic, mitral, tricuspid, or pulmonary valves is also a predictor of CVD. VC is associated with atherosclerosis, its risk factors, and its histopathologic profiles\cite{Rossi2021}. Two types of extracoronary VC are calcified aortic valve stenosis (CAVS) and mitral annular calcification (MAC)\cite{Shekar2018}. Although CT-based MAC severity scores exist, most are used to measure outcomes in valve embolization during transcatheter mitral valve replacement and mitral valve dysfunction rather than plaque burden\cite{Guerrero2020,Xu2023}.

Imaging modalities for cardiac plaque assessment include CT, MRI, and ultrasound\cite{Summers2021}.

While the majority of automated plaque detection models are typically trained using CT angiography scans, non-contrast CT scans are more prevalent because it is often easier to find plaques using these scans. One can readily identify calcification within coronary arteries by exploiting tissue density variations.\cite{Chen2020}. The lack of insurance reimbursement for CAC screenings makes automated opportunistic testing critical for potentially reducing cardiovascular disease\cite{Sandhu2023}. The purpose of this study is to validate an automated cardiac plaque detection model for non-contrast chest CT volumes.

   \begin{figure} [ht]
   \begin{center}
   \begin{tabular}{c} 
   \includegraphics[page=9, height=8cm]{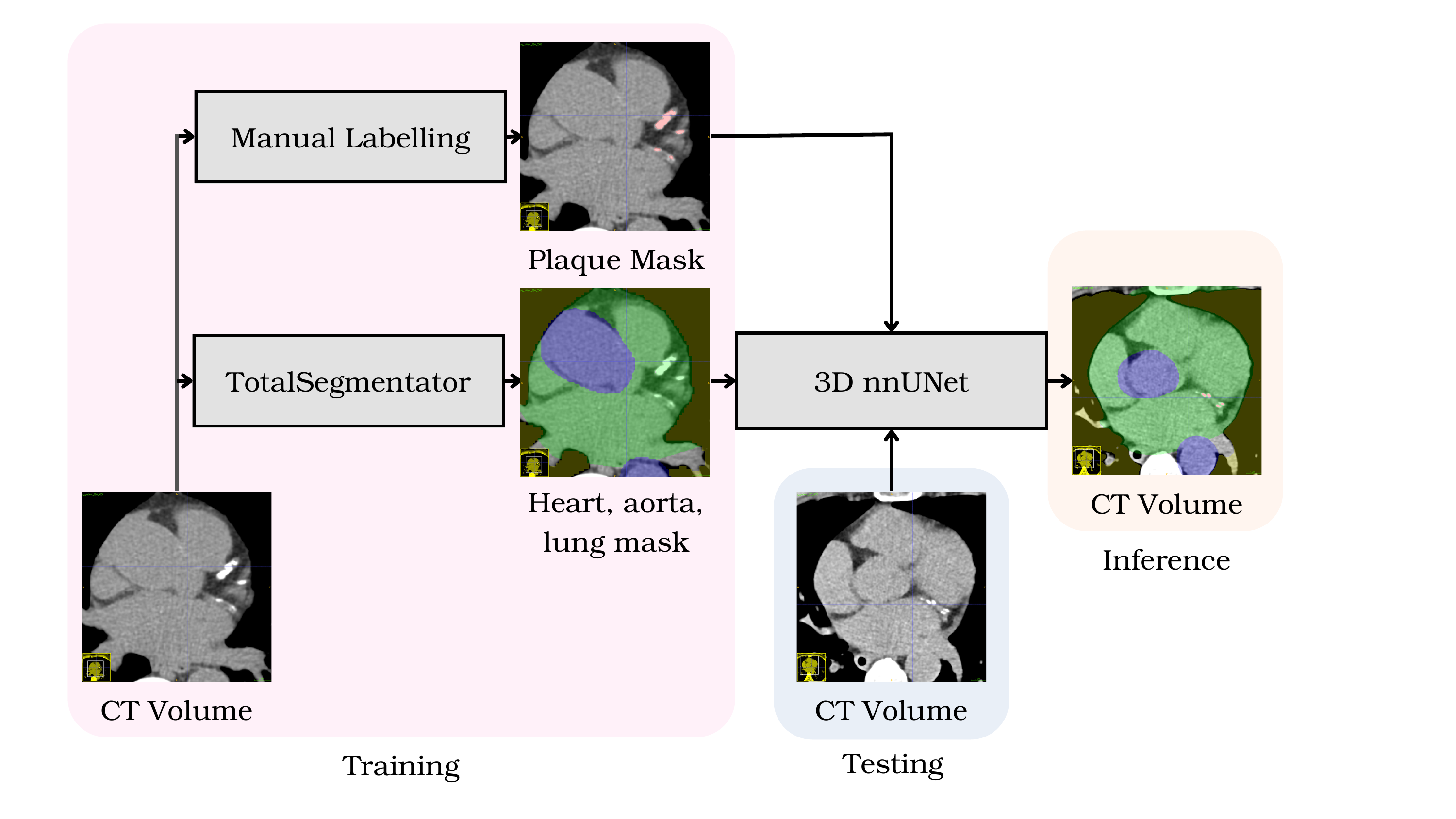}
   \end{tabular}
   \end{center}
   \caption[example] 
   { \label{fig:Sankey} 
Overall framework of the cardiac plaque detection algorithm. Given non-contrast CT scans, manual labelling segments plaques, while TotalSegmentator segments heart, aorta, and lung classes. Then, nnUNet generates 3D predictions from testing volumes.}
   \end{figure} 

\section{METHODS}

\subsection{Patient Population and Study Protocols}
\label{sec:title}

\begin{figure} [ht]
   \begin{center}
   \begin{tabular}{c} 
   \includegraphics[page=3, height=8
   cm]{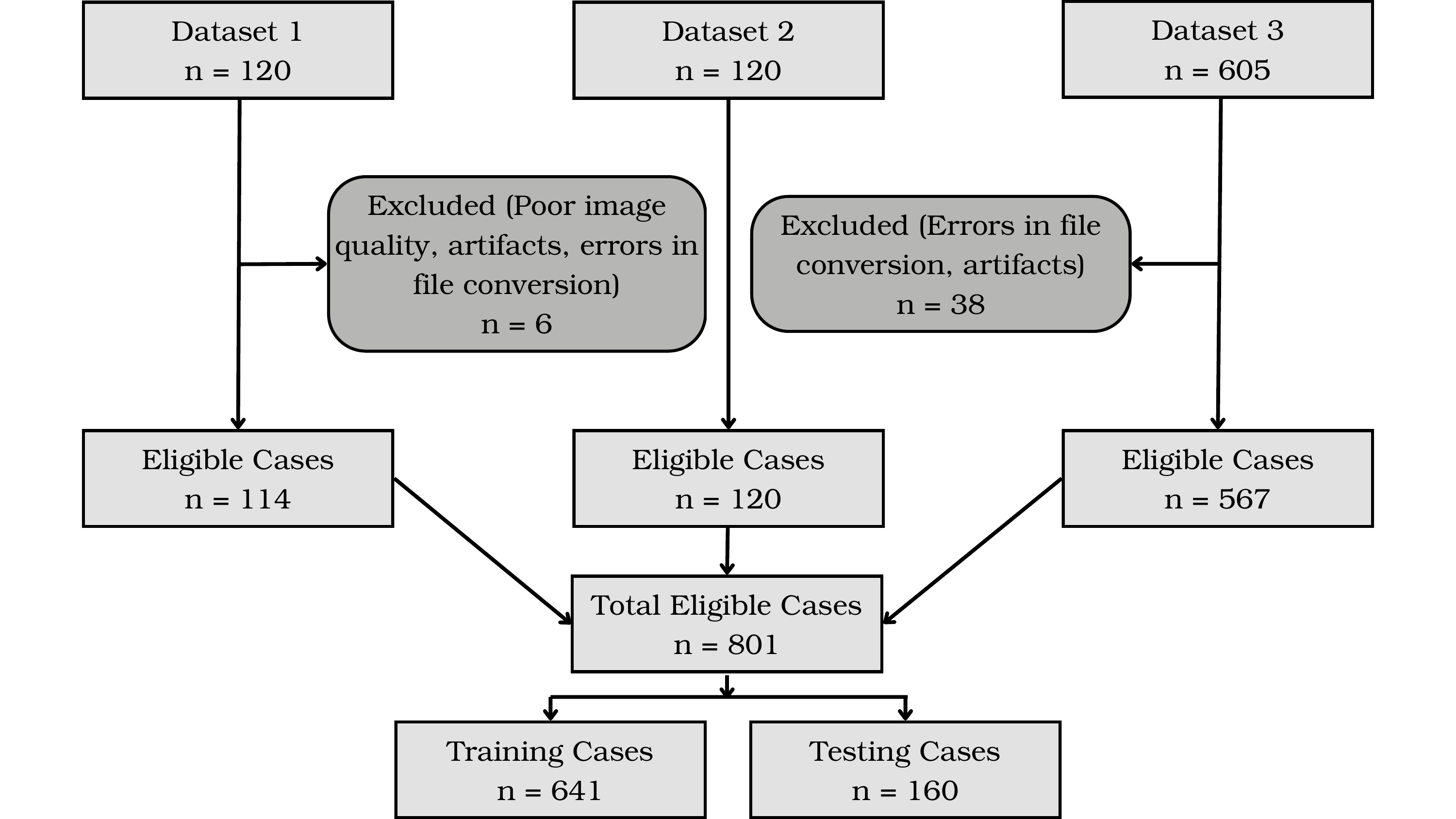}
   \end{tabular}
   \end{center}
   \caption[example] 
   { \label{fig:STARD} 
STARD Chart showing patient flow. In this graph, \( n \) represents the number of scans.}
   \end{figure}

CT scans were performed at three tertiary care hospitals and collected as three datasets, respectively. Cohort 1 is a consecutive series of 120 chest PET-CT scans of patients with vasculitis in a clinical trial at Institution 1\cite{Summers2021}. 6 volumes were excluded due to poor image quality, artifacts, or errors in conversion. Cohort 2 is a consecutive series of 120 non-contrast chest CT scans of patients diagnosed with CAD at Institution 2\cite{Kazemi2019}. Cohort 3 is a consecutive series of 605 non-contrast chest CT scans (392 gated and 213 non-gated) at Institution 3\cite{StanfordAIMI2021}. 38 volumes were excluded due to errors in file conversion and artifacts. In total, 801 total volumes were included and randomly split into a training set of 641 and a testing set of 160. A STARD chart showing exclusions and image assignment to training and testing sets is shown in Figure \ref{fig:STARD}.

\subsection{Data Preparation}
CT scans were converted from DICOM format into NIFTI volumes. Scans were reformatted to 3 mm slices and information on DICOM headers were extracted. NIFTI ground truth labels were available for datasets 1 and 2. XML files with ground truth plaque coordinates from dataset 3 were converted into NIFTI segmentations. All scans were reviewed and manually revised for missing plaque segmentations. Only axial CT volumes were assessed. Heart, aorta, and lung segmentations were determined using TotalSegmentator \cite{Wasserthal2022}. Segmentations were combined using add, dilation, and negation SimpleITK filter packages as well as convex hull operations\cite{Beare2018}. For all datasets, A.N. manually reviewed and labeled plaques in the coronary arteries and cardiac valves using ITK-Snap\cite{Yushkevich2016} under the mentorship of a board-certified radiologist (R.S., a radiologist with 30 years of post-residency experience reading chest CT).

\subsection{Cardiac Plaque Detection Algorithm}
We employed the 3D full resolution nnU-Net semantic segmentation pipeline to train the segmentation of plaque, heart, aorta, and lung classes\cite{Isensee2021}. Data preprocessing included cropping and image normalization. We used the default architecture of nnU-Net with Leaky ReLU nonlinearity. For the training process, we followed a 5-fold validation process with each fold running for 1000 epochs, where each epoch consisted of 250 iterations. The batch size was set to 2, optimizer was Adam, and the learning rate was set to 0.01. The loss function was a combination of cross-entropy and Dice loss.

\subsection{Statistical Analysis}
Precision and recall were calculated based on true positives (predicted plaques overlapping with ground truth), false positives (predicted plaques without ground truth overlap), and false negatives (ground truth plaques without predicted overlap). We performed linear regression to compare manual and automated Agatston scores across all datasets. The predicted Agatston score was computed using methods from Summers et al., considering plaque area, density, and associated weighting factors\cite{Summers2021}.

\section{Results}
\label{sec:sections}

Plaque detection was successful in 801 out of 845 (94.5\%) scans. The threshold for the detection was set at greater than two voxels. Detection precision was 0.893 and recall was 0.891. The total number of plaques in all datasets was 935, of which 839 were detected. The average Dice coefficient for the detected plaques was 0.75±0.16.

The most common cause of false positive detections in the test set was the misidentification of artifacts in the left anterior descending arteries and diagonal coronary arteries. These artifacts, often small, faint, and non-plaque in nature, were particularly prevalent in non-gated volumes with associated cardiac motion. The most common cause of false negative detections in the test set was the failure to detect smaller plaques in the right coronary artery and mitral valve.

\begin{figure} [ht]
   \begin{center}
   \begin{tabular}{c} 
   \includegraphics[page=4, height=8cm]{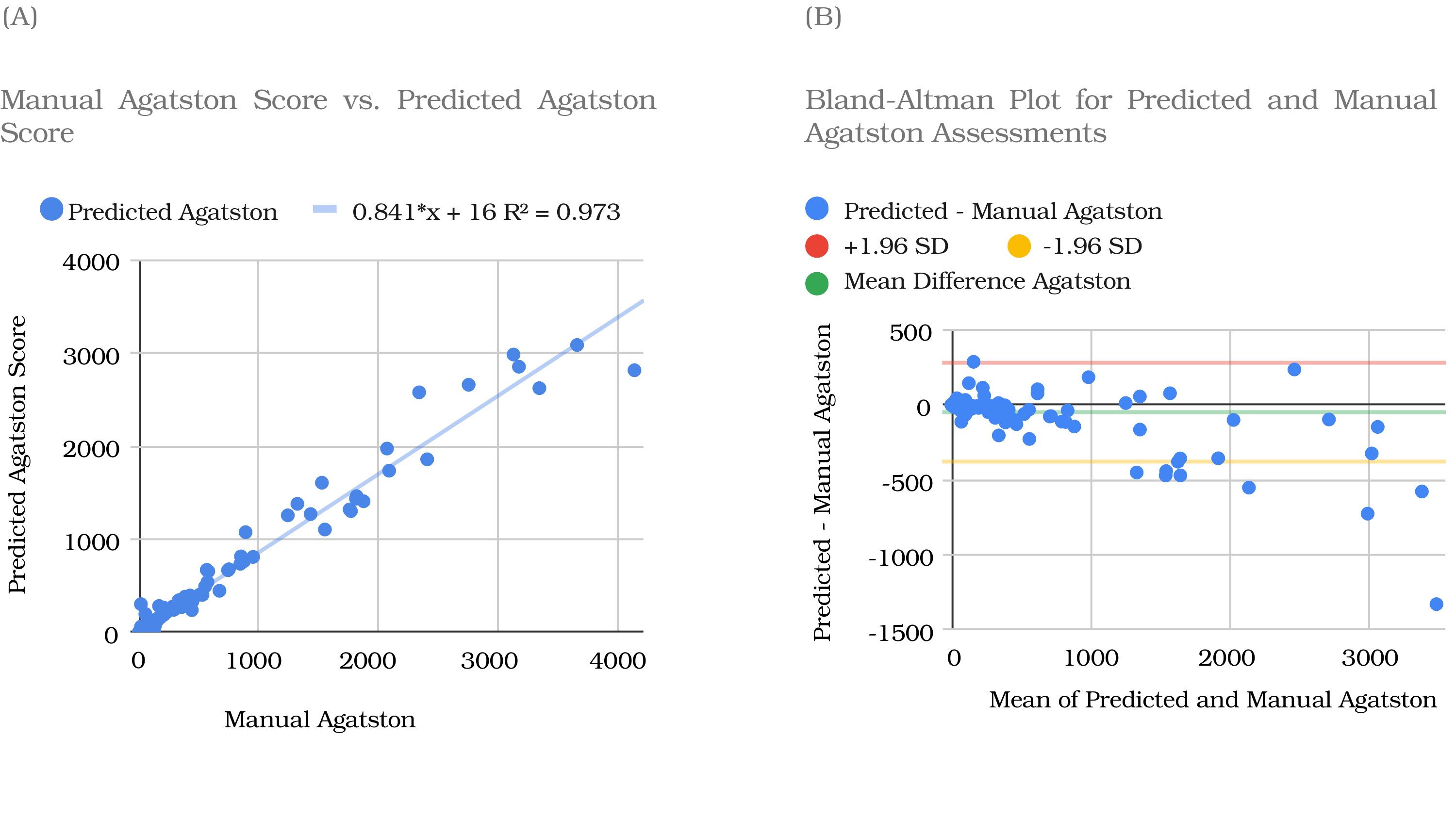}
   \end{tabular}
   \end{center}
   \caption[Agatston1] 
   { \label{fig:AgatstonAB} 
Agatston scores for the testing datasets. For the testing set, comparison of Agatston scores for automated and manual assessment showing (a) linear regression and (b) Bland-Altman plots.}
   \end{figure}

A scatterplot comparing manual and automated Agatston scores is shown in Figure \ref{fig:AgatstonAB}. The slope of the linear regression was 0.841 and the intercept was +16 HU (R² = 0.97). A Bland-Altman plot of the corresponding data is shown in Figure \ref{fig:AgatstonAB}.

\section{Discussion}
\label{sec:sections}

In this multicenter study, we adapt a multi-class 3D nnU-Net semantic segmentation pipeline to detect plaques in the coronary arteries and valves on non-contrast chest CT scans. By implementing a linear correction, the nnU-Net deep learning method provides precise estimation of Agatston scores on gated and non-gated scans. Notably, the correlation between manual and automated Agatston scores in this study demonstrates improvement over previous work assessing automated Agatston score computation in non-contrast non-gated CT scans (R² = 0.86)\cite{CanoEspinosa2018}.

Volumes with higher plaque burdens yielded greater disparities between predicted and manual Agatston scores, despite high precision and recall. This underestimation may be attributed to the presence of smaller plaques with more ambiguous areas compared to their larger counterparts. Numerous small plaques can amplify this underestimation, resulting in a divergence between predicted and manual Agatston scores.

False negatives were more commonly associated with the right coronary artery, likely owing to the relatively fewer volumes with plaques in that particular region in the training dataset. The disparity between left and right sides of the heart may also explain the underestimation in automated Agatston scoring. The detection of plaques in the valves emerged as a particularly challenging task. The complexity of plaque composition and distribution may impact the performance of the automated scoring system. This underscores the need for further research and validation, especially concerning the identification of smaller plaques within the right coronary artery and heart valves.

\section{Conclusion}
\label{sec:sections}

In this work we demonstrate how a multi-class 3D nnU-Net semantic segmentation pipeline and TotalSegmentator may be used in tandem to detect plaques in the coronary arteries and valves for non-contrast chest CT scans. We validated the performance of this plaque detector using three multicenter datasets and showed that it can accurately estimate Agatston scores on gated and non-gated scans.

\section{Acknowledgements}
\label{sec:sections}

This work was supported by the Intramural Research Program of the National Institutes of Health, Clinical Center and the NIH HPC Biowulf cluster.

\section{Appendix}
\label{sec:sections}
\appendix    

      \begin{figure} [ht]
   \begin{center}
   \begin{tabular}{c} 
   \includegraphics[page=7, height=8cm]{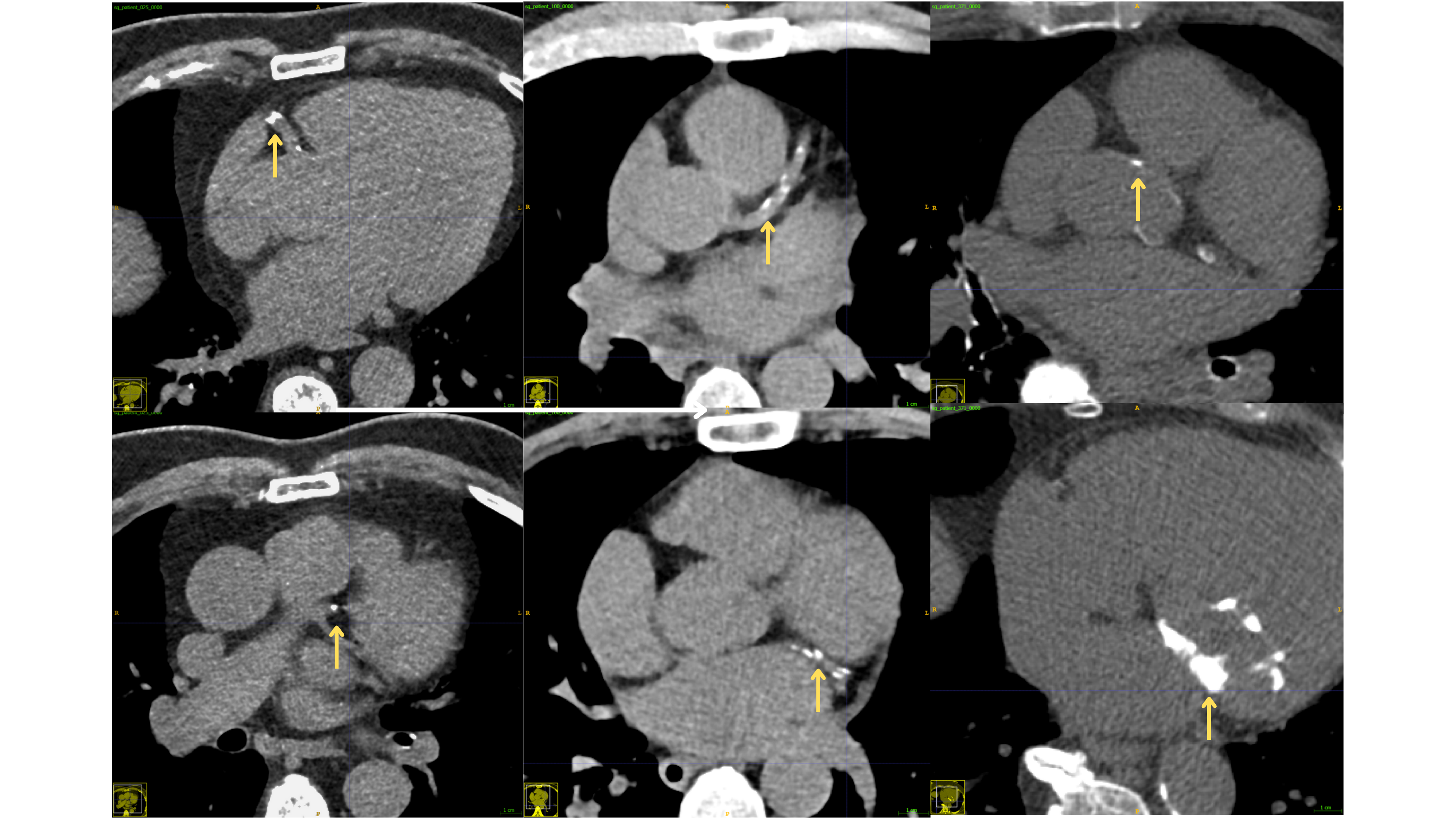}
   \end{tabular}
   \end{center}
   \caption[example] 
   { \label{fig:Plaque1} 
Examples of plaques (yellow arrows) in the coronary arteries and mitral valve in axial CT images.}
   \end{figure}

      \begin{figure} [ht]
   \begin{center}
   \begin{tabular}{c} 
   \includegraphics[page=8, height=8cm]{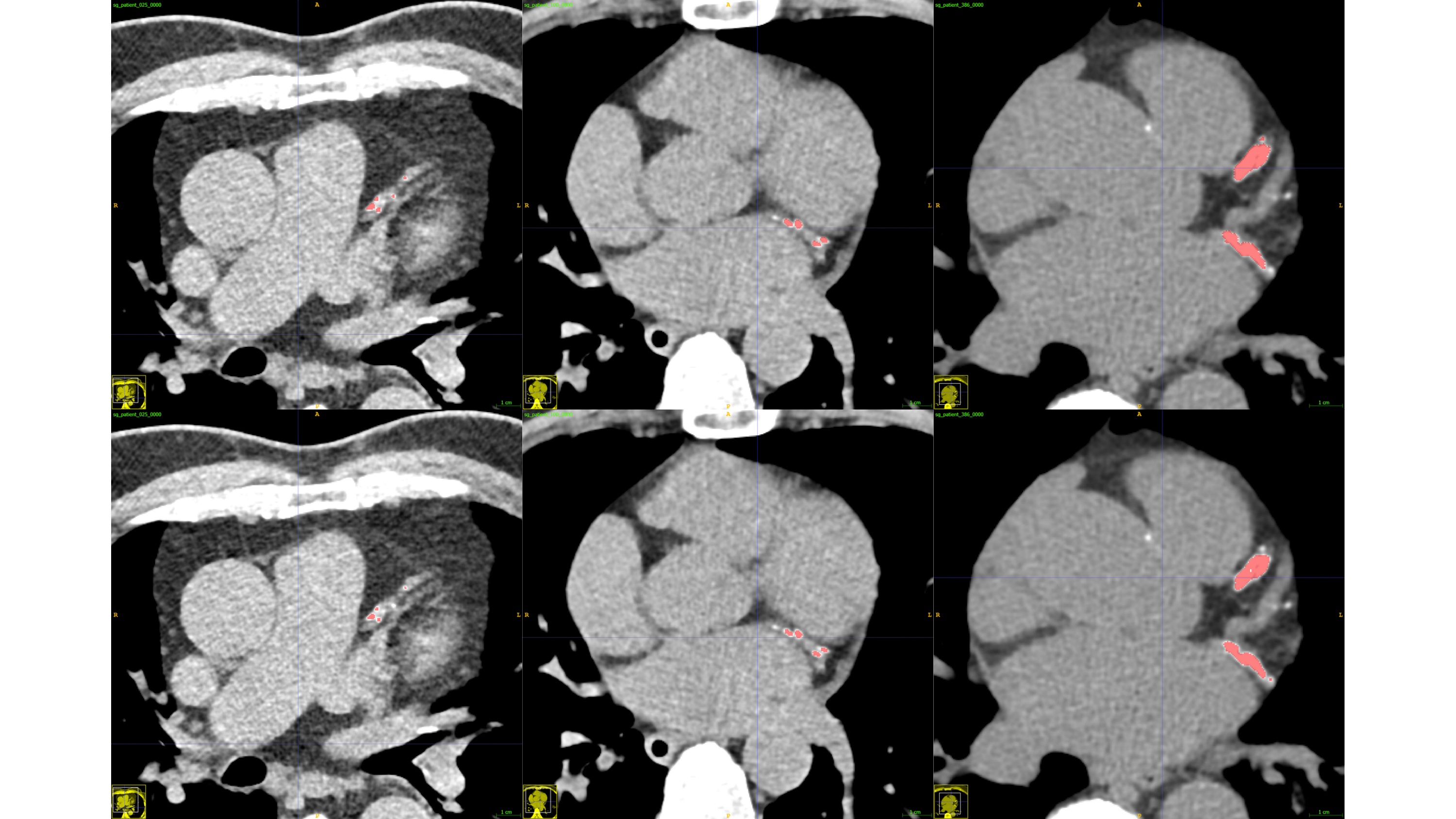}
   \end{tabular}
   \end{center}
   \caption[example] 
   { \label{fig:Plaque2} 
Examples of ground truth (top row) and corresponding predicted (bottom row) plaque burden measurements in axial CT images. For each example, the images are shown with labels and detections (red).}
   \end{figure}

\clearpage
\bibliography{report} 

\end{document}